\documentclass[10pt,twocolumn,letterpaper]{article}

\usepackage{iccv}
\usepackage{times}
\usepackage{epsfig}
\usepackage{graphicx}
\usepackage{amsmath}
\usepackage{amssymb}

\usepackage{color, colortbl}
\usepackage{booktabs}
\usepackage{rotating}
\usepackage{multirow}
\usepackage{enumitem}
\usepackage{sidecap}
\usepackage{xfrac}
\usepackage{etoolbox,siunitx}
\usepackage{booktabs}

\usepackage{amsmath}
\usepackage{amssymb}
\usepackage{sidecap}

\usepackage{graphicx}
\usepackage{caption}

\newcommand{\figref}[1]{Fig.~\ref{#1}}

\newcommand{\tabref}[1]{Table~\ref{#1}}

\newcommand{\RemoveAboveCaption}{-5pt}
\newcommand{\RemoveBelowCaption}{-10pt}
\newcommand{\RemoveAboveCaptiontab}{3pt}
\newcommand{\RemoveBelowCaptiontab}{-12pt}

\usepackage[breaklinks=true,bookmarks=false]{hyperref}

\iccvfinalcopy 

\def\iccvPaperID{****} 

\begin{document}

\title{6D Object Pose Estimation with Depth Images:\\A Seamless Approach for Robotic Interaction and Augmented Reality}

\author{David Joseph Tan, Nassir Navab, Federico Tombari\\
Technische Universit\"at M\"unchen
}


\twocolumn[{%
\renewcommand\twocolumn[1][]{#1}%
\maketitle

\vspace{-25pt}
\begin{center}
    \centering
    \includegraphics[width=\textwidth]{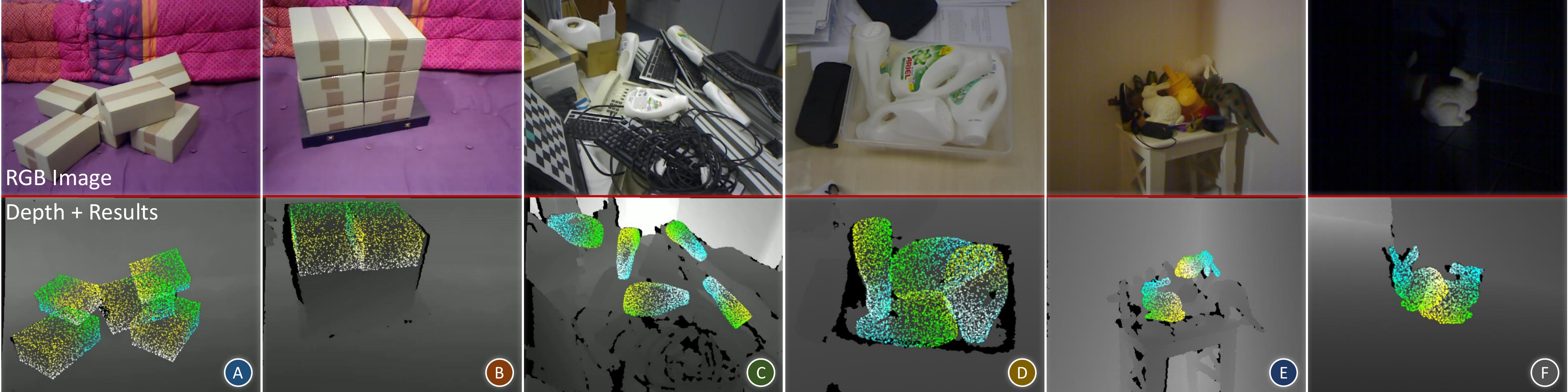}
\vspace{-20pt}
    \captionof{figure}{Results of the framework on different types of objects -- from simple to complex geometric structure -- on challenging scenarios.}
\label{fig:teaser}
\end{center}%
}]

\begin{abstract}
   To determine the 3D orientation and 3D location of objects in the surroundings of a camera mounted on a robot or mobile device,
we developed two powerful algorithms in object detection and temporal tracking that are combined seamlessly for robotic perception and interaction as well as Augmented Reality (AR).
A separate evaluation of, respectively, the object detection and the temporal tracker demonstrates the important stride in research as well as the impact on industrial robotic applications and AR.
When evaluated on a standard dataset, the detector produced the highest f1-score with a large margin while the tracker generated the best accuracy at a very low latency of approximately 2~ms per frame with one CPU core -- both algorithms outperforming the state of the art. 
When combined, we achieve a powerful framework that is robust to handle multiple instances of the same object under occlusion and clutter while attaining real-time performance. Aiming at stepping beyond the simple scenarios used by current systems, often constrained by having a single object in absence of clutter, averting to touch the object to prevent close-range partial occlusion, selecting brightly colored objects to easily segment them individually or assuming that the object has simple geometric structure, we demonstrate the capacity to handle challenging cases under clutter, partial occlusion and varying lighting conditions with objects of different shapes and sizes.
\end{abstract}

\section{Introduction}

The innovations in robotics have undergone numerous improvements to maneuver over rough terrains, to avoid collisions and even to take flight. However, a huge limitation for these autonomous agents is still represented by the lack of perception of the world around them. Without perception, modern robots are incapable of autonomously interacting with the objects present in their surroundings. Hence, although the advances in hardware to stabilize the robot's movement have reached remarkable achievements, the perception software still falls behind.

Most robotic perception tasks aim at providing an autonomous agent with the skill of either localizing itself in the surrounding environment, or interacting with nearby objects through grasping and manipulation. Our work has investigated on the latter skill, where we develop robotic perception techniques for real-time object detection and tracking of multiple 3D objects from RGB-D data acquired using consumer depth cameras. In particular, we have focused on industrial robotic applications, where intelligent robots have to localize, grasp, assemble and relocate objects on the production line, for tasks such as pick-and-place and bin picking. Looking from a wider perspective, the solution we developed is fundamental for most of the envisioned applications concerning service and personal robotics, where robotic assistants help out people for daily tasks in their domestic environments.
In addition, it is also fundamental to AR applications, where the pose of several objects has to be efficiently estimated while the user interacts with them.

Our framework is inspired by the way humans interact with the objects around them.
To be able to interact with objects in our surrounding, we first need to localize each object in the scene, then keep track of it throughout the process.
Converting this intuitive idea into an algorithm involves a two-step procedure -- object detection and temporal tracking. These two methods work hand-in-hand such that the former perceives the object in the scene while the latter keeps track of the object's movement during interaction.

\section{Seamless Object Detection and Tracking}

The goal of the framework is to find the objects in the scene while estimating their orientation and location in the 3D space, then to continuously track them throughout the following frames.
Object detection performs a sliding window approach to simultaneously find the objects in the scene and estimate their pose. 
Taking the resulting pose from the detector as input, the tracker estimates the relative transformation of the object between two consecutive frames and temporally relays the pose from one frame to the next. 
Intended for autonomous robots, the object detection and tracking run automatically in order to detect whenever the object of interest is present and stop tracking when the object is no longer visible.

Given the 3D CAD model of the object, both algorithms learn random forests~\cite{breiman2001random} from multiple, synthetically rendered depth images acquired by positioning the camera around the model.
Motivated to perform different tasks, the detector predicts the rotation and translation parameters for each region of the sliding window while the tracker predicts the relative transformation between two consecutive frames~\cite{tan2014multi,tan2017looking,tan15versatile}. 

\definecolor{ColorFilter}{rgb}{0.87,0.92,0.97}
\definecolor{OurResults}{rgb}{1.0,0.98,0.92}
\definecolor{TableBorder}{rgb}{0.8,0.8,0.8}

%
%

\begin{table*} [h]
\begin{center}
\begin{tabular}{
c|
>{\centering\arraybackslash}p{1.5cm}|
>{\centering\arraybackslash}p{1.5cm}|
>{\centering\arraybackslash}p{1.5cm}|
>{\centering\arraybackslash}p{1.5cm}|
>{\centering\arraybackslash}p{1.5cm}|
>{\centering\arraybackslash}p{1.5cm}|
>{\centering\arraybackslash}p{1.25cm}
}
\toprule
\multicolumn{1}{c}{} & 
\multicolumn{1}{c}{Coffee Cup} & 
\multicolumn{1}{c}{Shampoo} & 
\multicolumn{1}{c}{Joystick} & 
\multicolumn{1}{c}{Camera} & 
\multicolumn{1}{c}{Juice Carton} & 
\multicolumn{1}{c|}{Milk} & 
\multicolumn{1}{c}{\textbf{Mean}} \\
\midrule

LineMod~\cite{hinterstoisser2012accv}	&	94.2\% & 	92.2\% & 	84.6\% & 	58.9\% & 	59.5\% & 	55.8\% & 	74.2\% \\ 
Point-Pair Features~\cite{Drost2010Model}	&	86.7\% & 	65.1\% & 	27.7\% & 	40.7\% & 	60.4\% & 	25.9\% & 	51.1\% \\
Coordinate Reg.~\cite{Brachmann2014}	&	91.2\% & 	82.4\% & 	75.9\% & 	69.1\% & 	89.7\% & 	47.6\% & 	75.9\% \\
Latent Forest~\cite{tejani2014latent}	&	89.1\% & 	79.2\% & 	54.9\% & 	39.4\% & 	88.3\% & 	39.7\% & 	65.1\% \\
Next-Best-View~\cite{doumanoglou2016recovering}	&	93.2\% & 	73.5\% & 	92.4\% & 	90.3\% & 	81.9\% & 	51.0\% & 	80.4\% \\
Deep Learning~\cite{kehl2016deep}	&	97.2\% & 	91.0\% & 	89.2\% & 	38.3\% & 	86.6\% & 	46.3\% & 	74.8\% \\

\midrule                                     
\textbf{Our Detector}    &  99.8\% & 	99.2\% & 	98.9\% & 	99.0\% & 	99.7\% & 	99.3\% & 	99.3\% \\

\bottomrule
\end{tabular}
   \setlength{\abovecaptionskip}{\RemoveAboveCaptiontab}
   \setlength{\belowcaptionskip}{\RemoveBelowCaptiontab}
\caption{Comparison of f1-scores from object detection with 6D pose estimation algorithms, evaluated on the dataset of \cite{tejani2014latent}.}
\label{tab:detector}
\end{center}
\end{table*}

\definecolor{ColorFilter}{rgb}{1.0,0.95,0.8}
\definecolor{OurResults}{rgb}{0.93,0.93,0.93}
\definecolor{TableBorder}{rgb}{0.8,0.8,0.8}

\begin{table*} [!ht]
\begin{center}
\begin{tabular}{
c|
c|
>{\centering\arraybackslash}p{1.5cm}|
>{\centering\arraybackslash}p{1.5cm}|
>{\centering\arraybackslash}p{1.5cm}|
>{\columncolor{OurResults}}>{\centering\arraybackslash}p{2.0cm}|
>{\columncolor{ColorFilter}}>{\centering\arraybackslash}p{2.0cm}
}
\toprule 
	\multicolumn{5}{c}{} & 
	\multicolumn{2}{c}{\textbf{Our Tracker}} \\	
\cmidrule{6-7} 
	\multicolumn{1}{c}{} &
	\multicolumn{1}{c}{\emph{Errors}} &
	\multicolumn{1}{c}{PCL~\cite{rusu20113d}} &
	\multicolumn{1}{c}{C\&C~\cite{choi2013rgb}} &
	\multicolumn{1}{c}{Krull~\cite{krull6}} &
	\multicolumn{1}{c}{\emph{Learner}~\cite{tan15versatile}} &
	\multicolumn{1}{c}{\emph{AR}~\cite{tan2017looking}} \\
\midrule

\multirow{7}{*}{{\rotatebox[origin=c]{90}{(a)~\emph{Kinect Box}}}}												
&	$t_x$	&	43.99	&	1.84	&	0.83	&	0.24	&	0.15	\\
&	$t_y$	&	42.51	&	2.23	&	1.67	&	0.29	&	0.19	\\
&	$t_z$	&	55.89	&	1.36	&	0.79	&	0.18	&	0.09	\\
&	\emph{Roll}	&	7.62	&	6.41	&	1.11	&	0.17	&	0.09	\\
&	\emph{Pitch}	&	1.87	&	0.76	&	0.55	&	0.21	&	0.06	\\
&	\emph{Yaw}	&	8.31	&	6.32	&	1.04	&	0.16	&	0.04	\\
&	\emph{Time}	&	4539.0	&	166.0	&	143.0	&	1.4	&	2.2	\\
												
\midrule												
\multirow{7}{*}{{\rotatebox[origin=c]{90}{(b)~\emph{Milk}}}}												
&	$t_x$	&	13.38	&	0.93	&	0.51	&	0.27	&	0.09	\\
&	$t_y$	&	31.45	&	1.94	&	1.27	&	0.25	&	0.11	\\
&	$t_z$	&	26.09	&	1.09	&	0.62	&	0.21	&	0.08	\\
&	\emph{Roll}	&	59.37	&	3.83	&	2.19	&	0.24	&	0.07	\\
&	\emph{Pitch}	&	19.58	&	1.41	&	1.44	&	0.33	&	0.09	\\
&	\emph{Yaw}	&	75.03	&	3.26	&	1.9	&	0.25	&	0.06	\\
&	\emph{Time}	&	2205.0	&	134.0	&	135.0	&	1.4	&	2.1	\\
												
\midrule												
\multirow{7}{*}{{\rotatebox[origin=c]{90}{(c)~\emph{Orange Juice}}}}												
&	$t_x$	&	2.53	&	0.96	&	0.52	&	0.22	&	0.11	\\
&	$t_y$	&	2.2	&	1.44	&	0.74	&	0.21	&	0.09	\\
&	$t_z$	&	1.91	&	1.17	&	0.63	&	0.18	&	0.09	\\
&	\emph{Roll}	&	85.81	&	1.32	&	1.28	&	0.2	&	0.08	\\
&	\emph{Pitch}	&	42.12	&	0.75	&	1.08	&	0.24	&	0.08	\\
&	\emph{Yaw}	&	46.37	&	1.39	&	1.2	&	0.19	&	0.08	\\
&	\emph{Time}	&	1637.0	&	117.0	&	129.0	&	1.5	&	2.2	\\
												
\midrule												
\multirow{7}{*}{{\rotatebox[origin=c]{90}{(d)~\emph{Tide}}}}												
&	$t_x$	&	1.46	&	0.83	&	0.69	&	0.24	&	0.08	\\
&	$t_y$	&	2.25	&	1.37	&	0.81	&	0.24	&	0.09	\\
&	$t_z$	&	0.92	&	1.2	&	0.81	&	0.17	&	0.07	\\
&	\emph{Roll}	&	5.15	&	1.78	&	2.1	&	0.16	&	0.05	\\
&	\emph{Pitch}	&	2.13	&	1.09	&	1.38	&	0.3	&	0.12	\\
&	\emph{Yaw}	&	2.98	&	1.13	&	1.27	&	0.19	&	0.05	\\
&	\emph{Time}	&	2762.0	&	111.0	&	116.0	&	1.4	&	2.2	\\
												
\midrule												
\multirow{3}{*}{{\rotatebox[origin=c]{90}{\textbf{Mean}}}}												
&	Transl.	&	18.72	&	1.36	&	0.82	&	0.22	&	0.1	\\
&	Rot.	&	29.7	&	2.45	&	1.38	&	0.22	&	0.07	\\
&	Time	&	2786.0	&	132.0	&	131.0	&	1.4	&	2.2	\\
\midrule												
&	Hardware	&	CPU	&	\multirow{2}{*}{GPU}	&	\multirow{2}{*}{GPU}	&	CPU	&	CPU	\\
&	Requirement	&	(4 cores)	&		&		&	(1 core)	&	(1 core)	\\

\bottomrule
\end{tabular}

   \setlength{\abovecaptionskip}{\RemoveAboveCaptiontab}
   \setlength{\belowcaptionskip}{\RemoveBelowCaptiontab}
\caption{Comparison of the errors in translation (mm) and rotation (degrees), the failure rate (\%) and the runtime (ms) of the tracking results, evaluated on the dataset of \cite{choi2013rgb}.}

\label{tab:tracker}

\end{center}
\end{table*}

\section{Evaluation}

The framework satisfies various characteristics that are required by applications in robotics and AR. These include (1)~the robustness and accuracy to find the object in the scene and to estimate its pose; (2)~the efficiency to run in real-time with a minimal computational expense; and, (3) the cost-effective system requirements.

\subsection{Robust and Accurate to Detect and Track}

Fundamentally, the goal is to develop two powerful algorithms such that, in the combined approach, each algorithm can perform their assigned tasks very well.

\paragraph{Robust Detector with 6D Pose Estimate.}
Evaluated on the public dataset of \cite{tejani2014latent}, the object detection algorithm acquires the best f1-score in \tabref{tab:detector} with a large margin of 18.9\% against other methods~\cite{Brachmann2014,doumanoglou2016recovering,Drost2010Model,hinterstoisser2012accv,kehl2016deep,tejani2014latent}. We remind the reader that the f1-score measures the detection rate by incorporating both the precision and recall into one value. An f1-score of 99.3\% then indicates that the detection rate is almost perfect. 

\begin{figure} [!t]
\begin{center}
\includegraphics[width=\linewidth]{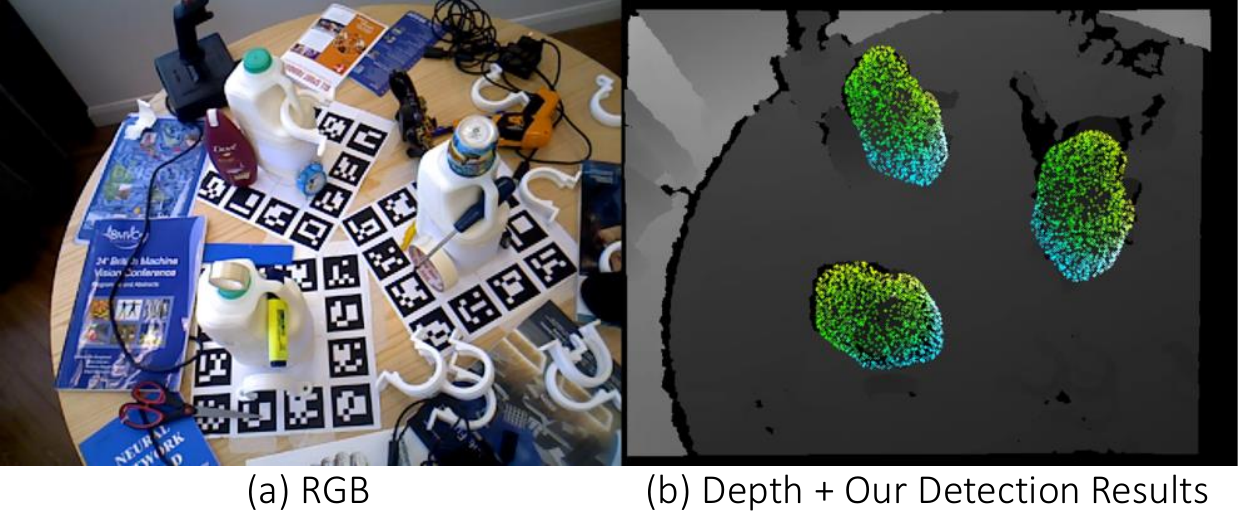}
\end{center}
   \setlength{\abovecaptionskip}{\RemoveAboveCaption}
   \setlength{\belowcaptionskip}{\RemoveBelowCaption}
   \caption{Sample RGB-D frame of the ``Milk'' sequence from \cite{tejani2014latent}.}
\label{fig:sample_frame_tejani}
\end{figure}

An interesting observation from \tabref{tab:detector} is that all the other methods~\cite{Brachmann2014,doumanoglou2016recovering,Drost2010Model,hinterstoisser2012accv,kehl2016deep,tejani2014latent} have a low f1-score on the ``Milk'' sequence. Among them, the highest reached 55.8\% from LineMod~\cite{hinterstoisser2012accv}.
The reason for the low results is because the authors from \cite{tejani2014latent} added small objects on the object of interest as shown in \figref{fig:sample_frame_tejani}. In effect, this occludes several regions of the object of interest and slightly changes its geometry. Contrary to their results, we can handle these occlusions and achieve an f1-score of 99.3\% on this sequence which is 43.5\% higher than the other methods.

\paragraph{Accurate Temporal Tracker.}
On the other hand, the temporal tracker generates the lowest errors in translation and rotation in \tabref{tab:tracker} on the dataset of \cite{choi2013rgb}. 
In addition, we also introduce a version of the tracker for AR~\cite{tan2017looking}.
Compared to \cite{tan2014multi,tan15versatile}, the latest version of the temporal tracker~\cite{tan2017looking} aims at the user experience for AR. Thus, it minimizes jitter by optimizing through the RGB and depth images to acquire more accurate pose estimations as evaluated in \tabref{tab:tracker}.

\paragraph{Combined Performance.}
Hence, the combined approach incorporates high detection rates as well as highly accurate pose.
Moving past the public datasets, \figref{fig:teaser} demonstrates the robustness of our algorithm in different types of challenging scenarios, including clutter, partical occlusions, and varying lighting conditions.
%
From simple to complex geometrical shapes, the results from \tabref{tab:tracker} and \figref{fig:teaser} show the capability of the framework to generalize its performance for different object shapes and sizes.

\begin{table} [!t]
\centering
\begin{tabular}{
c
!{\vrule width 0.6pt}
c
!{\vrule width 0.6pt}
c
}
\toprule
	\multicolumn{1}{>{\centering\arraybackslash}p{1.5cm}!{\vrule width 0.6pt}}{\emph{}} &
	\multicolumn{1}{>{\centering\arraybackslash}p{2.25cm}!{\vrule width 0.6pt}}{\textbf{Our Detector}} &
	\multicolumn{1}{>{\centering\arraybackslash}p{2.25cm}}{\textbf{Our Tracker}} \\
\midrule

Time	&	872.1~ms	&	1.4-2.2~ms	\\

\midrule

Computational 	&	CPU	&	CPU 	\\
 Power	&	 (8 Cores)	&	 (1 Core)	\\

\bottomrule
\end{tabular}
   \setlength{\abovecaptionskip}{\RemoveAboveCaptiontab}
   \setlength{\belowcaptionskip}{\RemoveBelowCaptiontab}
   \caption{Efficiency of the object detection and temporal tracker.}
\label{tab:efficiency}
\end{table}

\subsection{Low Latency and Low Memory Consumption}

\tabref{tab:efficiency} summarizes the detection time and tracking time with respect to the computational power.
Note that, after detecting the object, only the temporal tracker keeps track of the object in the subsequent frames. Hence, the latency of the framework depends on the temporal tracker alone, which is approximately 2~ms per frame for each object with a single CPU core. 
This efficiency is a substantial improvement against other temporal trackers from \tabref{tab:tracker} that run at 2786.0~ms for \cite{rusu20113d}, 132.0~ms for \cite{choi2013rgb} and 131.0~ms for \cite{krull6}, where \cite{choi2013rgb,krull6} use GPU.
We further evaluated that we achieve a real-time performance in tracking 108 moving objects at 30 fps with 8 CPU cores~\cite{tan15versatile}. 
Moreover, considering that we have a learning-based approach that needs to store the random forests, we efficiently achieved a low memory footprint of about 40.5~MB per object.

\subsection{Low-Cost Hardware Requirements}

The hardware requirements for our framework is (1)~a standard computer;   
and, (2)~a consumer depth sensor (\eg Microsoft Kinect and Kinect II, Asus Xtion, PrimeSense Carmine, Orbbec Astra) that costs about 150~USD. 
These sensors are much cheaper than the industrial 3D sensors which are currently used.
In addition, due to the framework's efficiency, the cost remains low since powerful GPU's are not required because both the tracker and the detector only use the CPU cores. Here, all the quantitative and qualitative evaluations use an Intel(R) Core(TM) i7 CPU in a Lenovo W530 laptop.

\section{Conclusion}

We present a seamless object detection and tracking framework that automatically finds the object of interest in the scene and keeps track of these objects across time. 
Our evaluation proves that its robustness, accuracy, efficiency and cost-effective characteristics make it an ideal framework for applications in robotic perception and interaction for industrial robotics in the production line. 
Another set of applications is in Augmented Reality (AR), Mixed Reality (MR) and Virtual Reality (VR), where not only does the pose estimation play a fundamental role in finding the objects in the scene but it is also an enabling technology in a pipeline composed of multiple modules running on the same power- and memory-limited hardware platform.
Therefore, this allows applications to have ample of time to build or render on top of our framework as well as the capacity to fully utilize the machine's hardware resources.
Notably, it is also highly suitable for applications in mobile platforms.

\section{Videos}

We prepared demonstrative videos\footnote{Link to video: \url{https://youtu.be/7rKBZZHJkFk}}\footnote{Link to video: \url{https://youtu.be/1P184ZocMo8}} to show the framework's performance. 
For AR applications, we introduce another set of videos\footnote{Link to video: \url{https://youtu.be/t-WDIqEPQ3g}}\footnote{Link to video: \url{https://youtu.be/8-0xsc2abQs}} that uses our new temporal tracker \cite{tan2017looking} to estimate a more accurate pose and to handle challenging scenarios.


\vspace{10pt}

{\setlength{\parindent}{0cm}
\textit{
This extended abstract was submitted to the demo session of \textbf{ISMAR 2017} and the 3rd International Workshop on Recovering 6D Object Pose organized at \textbf{ICCV 2017}.
%
}}

{\small
\bibliographystyle{ieee}
\bibliography{string,vision}
}

\end{document}